# A Trilogy of AI Safety Frameworks:
## Paths from Facts and Knowledge Gaps to Reliable Predictions and New Knowledge

Simon Kasif
Department of Biomedical Engineering
Program in Bioinformatics
Department of Computer Science
Boston University

AI Safety has become a vital front-line concern of many scientists within and outside the AI community. There are many immediate and long term anticipated risks that range from existential risk to human existence to deep fakes and bias in machine learning systems [1-5]. In this paper, we reduce the full scope and immense complexity of AI safety concerns to a trilogy of three important but tractable opportunities for advances that have the short-term potential to improve AI safety and reliability without reducing AI innovation in critical domains. In this perspective, we discuss this vision based on several case studies that already produced proofs of concept in critical ML applications in biomedical science.

**Linking Predictions to Knowledge (P2K)**

The remarkable acceleration of AI technologies is effectively driving a flood of AI based methods and predictions. We are literally drowning in a sea of predictions and AI generated products and this trend is projected to increase. The realization accentuates the necessity of factual grounding and the importance of experimentally validated and transparently documented facts. This is perhaps best highlighted by a famous quote from a legendary scientist cited in numerous publications (e.g. [5]). "False facts are highly injurious to the progress of science, for they often endure long; but false views, if supported by some evidence, do little harm, for everyone takes a salutary pleasure in proving their falseness; and when this is done, one path towards error is closed and the road to truth is often at the same time opened." – Charles Darwin

Following on these, widely accepted observations, we will argue that AI/ML systems must invest a significant effort and resources to develop transparent paths from predictions to facts. We use the term P2K (Predictions to Knowledge) to describe provenance or validation paths from predictions back to validated facts and established knowledge. This grounding is an essential step in converting knowledge, data and predictions into new knowledge. In mathematics such paths are simply called proofs. In AI based biological science P2K can take many forms that collectively generate an easy to validate factual foundation and provenance linking predictions to well documented facts. As one early example, the COMBREX (Computational Bridges to Experiments) project [6] is an implementation of the (2AI2BIO)* paradigm, namely a repeated loop that drives biological research using AI. COMBREX integrated AI in the form of Active Learning and Amazon Turk models to catalyze a community effort aimed at Gene Function characterization. A primary focus was on experimental testing of gene function predictions produced by ML software, thereby increasing the number of facts that we referred to as "gold standard annotation" essential for accurate prediction. The COMBREX project prioritized experimental testing using information gain and other criteria popularized in AI and Active Learning but not previously deployed systematically to catalyze community science in biology. Such prioritized experiments increase the accuracy of overall AI predictions globally, beyond a validation of a specific prediction or a curiosity driven hypothesis. However, COMBREX also introduced a rudimentary P2K method to trace predictions to their putative experimental sources (facts). For each gene X has a biochemical function Y prediction, COMBREX identified the most similar experimentally characterized genes with fully documented functional annotation, that may or may not support the predicted annotations supplied [5,6]. More generally, P2K "proofs" may rely on the specific ML, Generative AI or other methods used to make predictions or produce solutions in the form of text, images, plans, structures (e.g., molecules) or other representations. For machine learning experts among the readers, the AI systems may generate a relevant kernel (distance function induced by the ML method with rigorous mathematical properties) [10]) to "connect" the predictions to each other and their "closest" experimentally validated sources. Efficient methods linking predictions to facts may use methodologies such as graph kernels and diffusion to produce P2K paths [7].

Based on this philosophy, a simple P2K method was developed as part of an effort aimed at identifying new targets for drug repositioning for COVID [7]. In particular, predicted drug targets were traced to human proteins that were experimentally shown to interact with the virus. The P2K trace in this case was implemented using an efficient network propagation procedure analyzing the trace of a network diffusion [7] to identify and report the experimental source of each prediction. More recently, the important area of "factual grounding" that emerged as a response to hallucinations in language models is related to P2K and accentuates and reinforces the ethical, scientific and AI safety needs for gold standard and ground truth. P2K grounding procedures can be devised for any knowledge graph or a guilt by association network that are used by many communities.

These "fact finding/tracing" P2K methods are related but different from the more widely used explanation-based methods in interpretable AI systems. Explanations are not necessarily traced to ground truth. Thus, we must distinguish between experimentally determined and fully documented hard facts and explanations that more often than not use known or latent features to explain predictions. For instance, an AI system that predicts an enzyme to have a metabolic function F, might use a reference to a neighbor in bacterial genome has the same predicted function (with no facts to support it). In COMBREX, **factual grounding** of a gene function prediction required a trace to the experiment documenting the conditions, the organism and context where the biochemical function was established. In a different biomedical context, consider the case where a prediction of cancer risk can be attributed to a mutation in the BRCA2 gene. In some cases, provably pathogenic mutations are stored in highly curated databases of clinically and biologically documented cases. In most cases however, pathogenicity is predicted and associated with an explanation that may include some structural or functional property of the mutated residue suggesting pathogenicity. E.g., a mutation of a residue that is relatively conserved across functionally related proteins or occurs on a protein surface (thus more likely to interact with the substrate the protein acts on). These explanations are not facts. They are just evidence that reduces the uncertainty associated with the prediction. A strict P2K provenance would require providing links to concrete experimental evidence that proves that the mutation is causing a disruption (e.g. loss of function via an experimental screen or biochemical testing) and/or documentation of the clinical impact in patients or patient families where this specific mutation was observed. While provenance, probabilistic or otherwise, is challenging to establish, our overall vision for safe and reliable AI systems includes the platforms for P2K links tracing predictions to facts (in addition to explanations). The legal system provides perhaps an easy to understand motivation and analogy of tracing an argument in a new legal case via **precedence** to a decided and well documented prior case.

**AI Inspector Systems (AIIS)**

Inspectors are generally found in literally every conceivable industry that deals with widely used products. From underwear (famously documented in commercials with celebrity athletes such as Michael Jordan) to airplanes and cars. Many AI and Data Scientists have proposed different forms of inspectors to **explain, challenge, validate or analyze the quality of AI generated predictions and representations**. However, "AI Inspectors" are largely missing from a systematic deployment in AI systems, despite their increasingly broad use. This slow deployment of inspectors is likely related to the relatively young age of widely deployed AI systems.  Given the countless AI risks, the industry should consider a major investment in the development and release of AI inspectors especially in critical domains such as science and medicine.

For length considerations, we would just allude to one inspector developed in conjunction to an ML system for predicting gene fusions in tumors [8].  Identifying gene fusions in cancer cells can lead to exceptionally effective drugs targeting the specific abnormal fusion of two genes in cancerous tissues. In one particularly striking and clinically important case, the drug Gleevec produces an almost perfect remission in tumors associated with the targeted gene fusion.  Predictive genomic ML systems are playing an increasingly important role in Precision Medicine.   However, these systems may have significant error rates that depend on a number of complex factors from basic biology to other engineering, platform or data science challenges.  The inspector method described in [9] deployed unsupervised learning to cluster all predictions into "uncertainty" classes. These clusters have different level of enrichment in experimental validations and consequently different error rates. The inspector method analyzes these clusters and uses features extracted from these classes to provide confidence estimates for gene fusion predictions based on their membership or association with these clusters.  Curiously, the inspector system implemented in this predictive genomic system is conceptually related to broader ML ideas related to quantifying uncertainty for each prediction.  Uncertainty quantification (UQ) [11-14] is a useful research area that consists of rigorous mathematical work, elegant visualization procedures and software support.

**Knowledge Gaps and Uncertainty Qualification:**

In addition to UQ we propose the need to learn the "ignorance" regions, we refer to as Knowledge Gaps (KGs) where we do not have a satisfactory number of hard facts and/or have a very high uncertainty in predictions.  In fact, the COMBREX project proposed to produce a searchable database of knowledge gaps [6,9] where ML systems cannot make reliable gene function predictions.  However, what is a Knowledge Gap?  We leave the formal definition to follow-up papers but intuitively it's a) a large subset of the domain where any predictions are associated with a very high uncertainty b) can be succinctly described or summarized using some representation, e.g. a logical formula, decision tree, ontology, a subnetwork, geometric manifold, bi-cluster, natural language, a small number of prototypes or others.   For instance, consider an AI system for a binary class prediction (such as a diagnosis $D(X) \rightarrow \{0,1\}$). A KG is the set of instances $\{X\}$ such that entropy of $P(D(X) = - P(D(X) = 1)\log (P(D(X) = 1)) - P(D(X) = 0)\log (P(D(X) = 0))$ is very high and ideally the KG will have a simple to interpret description that will allow both scientists or AI systems to investigate or explore effectively. Fig 1, illustrates KGs and P2K in three classical platforms for machine learning: a) 2D projection of data (with or without kernels), b) networks and provenance via diffusion on graphs and c) hierarchical provenance tracing in trees (e.g. in phylogenies).

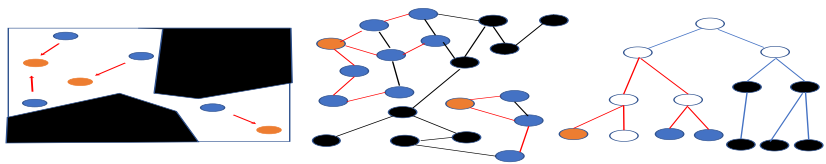

**Figure 1: Illustration of knowledge gaps (KG) and provenance tracing (P2K) in different predictive spaces. The nodes are data objects: e.g. genes or cells.  A) Real-valued 2D projections of data (e.g.  as in PCA or tSNE). B) Networks. C) Hierarchies (e.g. phylogenies).  Black nodes and regions = Knowledge Gaps, Gold nodes = Facts (e.g., genes associated with experimentally determined biochemical or biological function), Blue nodes = objects that have reliable predictions (e.g., genes associated with predicted functions).  Red edges are P2K paths linking predictions with facts.  This figure abstracts out the specific machine learning methods used.**

The utility and complexity of KGs is profound. Consider, the fundamental problem of annotating every gene in any human cell with a context specific function prediction. Contexts may include a specific tissue, condition (such as cancer, developmental stage or aging), environment. perturbation or treatment.  Function predictions may include biochemical function (e.g., DNA binding), biological function (e.g., DNA repair) or phenotype (e.g., increase risk for breast cancer in the case of BRCA1).  A different example would be the space of drugs, genes, diagnostics (tests), prognostics and diseases to drive precision medicine. At present, we have no catalogues of searchable and easy to understand knowledge gaps in any biomedical space that integrates experimental knowledge and our ability to make reliable predictions.  The search for KGs can be formalized as a computational identification of  large sub-spaces in the predictive space with a high mean uncertainty score and a relatively simple computable description of the KG.  KGs can be then be prioritized for new experimental studies and/or validations.  We will illustrate one example of KG search using a simplified algorithm below.

1. Generate hypotheses in the form of logical statements (or other representations such as natural language statements, images, web pages) by a probabilistic sampling procedure (as done by many generative AI systems).  For gene function prediction systems, the algorithm might sample the product space of genes and their putative function assignments or do a random walk on functional gene linkage networks [15].
2. For each prediction (e.g. gene G has function F) deploy UQ to associate it with an uncertainty score.
3. Use Reinforcement Learning (or other optimization) to identify large subspaces of the predictive space that have a relatively high average uncertainty scores and simple descriptions.  Various optimization methods that are common place in anomaly detection on networks can be extended and scaled up with AI ideas to characterize these KG regions.
4. Organize the KGs using established or novel methods into bi-clusters, connected communities in networks, ontologies, probabilistic models or other descriptive or searchable representations.  Visualize KGs using on tools such as PCA or tSNE.  Language models may be used to answer questions about KGs.

There are many AI areas relevant to our KG proposal. In particular, Active Learning (AL) devises trade-off between exploration (of the "unpredictable") and exploitation (testing high probability predictions).  In this paper, we argue for the need to develop methods for learning to predict, describe and/or search KGs that can drive scientific discovery.   In a different, yet relevant research area, AI scientists devised methods to model the continuous variation in prediction entropy over the full domain.  As a special case, identification of bias in machine learning systems, was studied in depth by many leading machine learning scientists [19]. A well-known example was the variation in classification accuracy across sub-populations creating racial disparities. However, in most of these studies, KGs were often described as well characterized sub-populations defined by a few variables such as race or locale.  We are seeking a more general definition of KGs for AI applications in scientific discovery and prioritization of experiments, where the boundaries or searchable features of KGs are yet to be known, thus must be learned.  We are seeking KG frameworks that allow AI systems to learn, predict, describe, search and understand the unpredictable [20].

**Human-centered exploration and validation**

We outlined a trilogy of three AI safety directions: P2K tracing of predictions to knowledge and facts, AI inspectors (AIIS) and knowledge gaps (KGs).  These directions should ideally catalyze human-centered and judiciously prioritized prediction-validation cycles. Building AI systems is not sufficient. AI Safety Systems should be complemented by a significant investment in resources and personnel allocated to factual validations in the Knowledge Gaps in the AI predictive space in a cost-effective manner, e.g. by Active Learning /Reinforcement Learning or other creative AI frameworks.  These resources should be in part directed towards establishing systematically prioritized and richly documented Ground Truth Databases of factual knowledge.  The validations might be carried by robotic pipelines, Citizen's Science or human Amazon Turk type platforms as attempted in COMBREX or more broadly by systematically organized and funded citizens science, community or commercial efforts [6,21,22]. Last but not least, we must also note that internal logical inconsistencies present significant challenges to building safe AI systems and require a serious impetus for introducing (at least) limited logical (verification or consistency) inference tools as part of the AI platform. Thus, in the long term, the integration of neural methods and symbolic proofs might be a very attractive approach for P2K and AIIS.

The trilogy described here is directly applicable to a very wide span of data science and AI methods used in biomedical sciences and beyond. The methods may use classical data science methods (e.g., as deployed in GWAS), guilt by association [15], causal methods [16], or modern deep neural networks or their integration with symbolic methods. However, AI safety is a serious concern for all methods given the growing range of applications and deployment in critical domains. While we described only a few demonstrations in biomedical sciences, the proposed trilogy is naturally applicable to many AI applications. Knowledge is indispensable and as noted by another legendary mind "An investment in knowledge pays the best interest". Benjamin Franklin. NIH, admirably, created the BD2K (Biological Data to Knowledge) initiative that aims to build on machine learning advances. However, knowledge is not equal to predictions, especially at scale produced by generative AI. Honestly, it is too easy to predict. Statistical validation is indeed important to prioritize experimental, clinical or mathematical validations. But facts, cannot be substituted with explanations and interpretations, thus P2K in the form of "grounding", provenance or tracing predictions or AI generated representations to facts is a useful feature of safe AI systems [7,17,18]. Smart AI inspector systems (AIIs) are indispensable future tools for quality control. Characterizing the boundaries/limits of AI systems to make reliable predictions, defined here as KGs is vital to safe AI. Finally, as we and others argued in the past, Human Centered AI (HCAI) leads to safer AI [18]. The trilogy proposed here can partially support reliable HCAI platforms to generate new and reliable knowledge.

**Acknowledgement**

The views expressed here are the author's alone. The work was influenced by collaboration, discussion, communication, encouragement or input from many scientists. We thank Richard J. Roberts, Martin Steffen, Yoshua Bengio, Stuart Russell, Aviv Regev, Charles Cantor, Charles DeLisi, Brian Haas, Adi Karni, Stan Letovsky, Robert Plenge, Judea Pearl, T.M. Murali, Mark Crovella, Jim Collins, Roded Sharan, Yishay Mansour, Dana Angluin, David Lipman, Eytan Ruppin, Jim Collins, Zak Kohane, Norbert Perrimon, Sebastian Seung, Ron Shamir and many others. We also specifically thank the leaders of the AI Safety movement. We also acknowledge NIH, NSF, Boston University, Harvard Medical School, MIT/Harvard Health Sciences and Technology Program, Children's Hospital, Joslin Diabetes Center, and other agencies and institutions support for the development of ideas and methods relevant to this paper. Part of this work was done when the author was visiting Tel Aviv University.

**Selected References**

1. Kurzweil, R., 2005. The singularity is near. In Ethics and emerging technologies (pp. 393-406). London: Palgrave Macmillan UK.

2. Bengio, Y., Hinton, G., Yao, A., Song, D., Abbeel, P., Darrell, T., Harari, Y.N., Zhang, Y.Q., Xue, L., Shalev-Shwartz, S. and Hadfield, G., 2024. Managing extreme AI risks amid rapid progress. Science, 384(6698), pp.842-845.

3. Cohen, M.K., Kolt, N., Bengio, Y., Hadfield, G.K. and Russell, S., 2024. Regulating advanced artificial agents. Science, 384(6691), pp.36-38.

4. Kasif, S., 2020. Artificial Tikkun Olam: AI Can Be Our Best Friend in Building an Open Human-Computer Society. arXiv preprint arXiv:2010.12015.

5. Kasif, S. and Roberts, R.J., 2020. We need to keep a reproducible trace of facts, predictions, and hypotheses from gene to function in the era of big data. PLoS Biology, 18(11), p.e3000999.

6. Anton, B.P., Chang, Y.C., Brown, P., Choi, H.P., Faller, L.L., Guleria, J., Hu, Z., Klitgord, N., Levy-Moonshine, A., Maksad, A. and Mazumdar, V., 2013. The COMBREX project: design, methodology, and initial results. PLoS biology, 11(8), p.e1001638.

7. Law, J.N., Akers, K., Tasnina, N., Santina, C.M.D., Deutsch, S., Kshirsagar, M., Klein-Seetharaman, J., Crovella, M., Rajagopalan, P., Kasif, S. and Murali, T.M., 2021. Interpretable network propagation with application to expanding the repertoire of human proteins that interact with SARS-CoV-2. GigaScience, 10(12), p.giab082.

8. Haas, B.J., Dobin, A., Ghandi, M., Van Arsdale, A., Tickle, T., Robinson, J.T., Gillani, R., Kasif, S. and Regev, A., 2023. Targeted in silico characterization of fusion transcripts in tumor and normal tissues via FusionInspector. Cell Reports Methods, 3(5).

9. Chang, Y.C., Hu, Z., Rachlin, J., Anton, B.P., Kasif, S., Roberts, R.J. and Steffen, M., 2016. COMBREX-DB: an experiment centered database of protein function: knowledge, predictions and knowledge gaps. Nucleic acids research, 44(D1), pp.D330-D335.

10. Hofmann, T., Schölkopf, B. and Smola, A.J., 2008. Kernel methods in machine learning.

11. Abdar, M., Pourpanah, F., Hussain, S., Rezazadegan, D., Liu, L., Ghavamzadeh, M., Fieguth, P., Cao, X., Khosravi, A., Acharya, U.R. and Makarenkov, V., 2021. A review of uncertainty quantification in deep learning: Techniques, applications and challenges. Information fusion, 76, pp.243-297.

12. Singh, G., Moncrieff, G., Venter, Z., Cawse-Nicholson, K., Slingsby, J. and Robinson, T.B., 2024. Uncertainty quantification for probabilistic machine learning in earth observation using conformal prediction. Scientific Reports, 14(1), p.16166.

13. Dolezal, J.M., Srisuwananukorn, A., Karpeyev, D., Ramesh, S., Kochanny, S., Cody, B., Mansfield, A.S., Rakshit, S., Bansal, R., Bois, M.C. and Bungum, A.O., 2022. Uncertainty-informed deep learning models enable high-confidence predictions for digital histopathology. Nature communications, 13(1), p.6572.

14. Nehme, E., Yair, O. and Michaeli, T., 2023. Uncertainty quantification via neural posterior principal components. Advances in Neural Information Processing Systems, 36, pp.37128-37141.

15. Murali, T.M., Wu, C.J. and Kasif, S., 2006. The art of gene function prediction. *Nature biotechnology*, *24*(12), pp.1474-1475.

16. Bareinboim, E. and Pearl, J., 2016. Causal inference and the data-fusion problem. Proceedings of the National Academy of Sciences, 113(27), pp.7345-7352.

17. Geva, M., Bastings, J., Filippova, K. and Globerson, A., 2023. Dissecting recall of factual associations in auto-regressive language models. arXiv preprint arXiv:2304.14767.

18. Huang, T., Jones, P. and Kasif, S., 1997. Human-Centered Systems: Information, Interactivity, and Intelligence. Report, NSF.

19. Kearns, M. and Roth, A., 2019. *The ethical algorithm: The science of socially aware algorithm design*. Oxford University Press.

20. Zabell, S. L. (1992) "Predicting the Unpredictable," Synthese 90, 205-232.

21. Cooper, S., Khatib, F., Treuille, A., Barbero, J., Lee, J., Beenen, M., Leaver-Fay, A., Baker, D. and Popović, Z., 2010. Predicting protein structures with a multiplayer online game. *Nature*, *466*(7307), pp.756-760.

22. Arganda-Carreras, I., Turaga, S.C., Berger, D.R., Cireşan, D., Giusti, A., Gambardella, L.M., Schmidhuber, J., Laptev, D., Dwivedi, S., Buhmann, J.M. and Liu, T., 2015. Crowdsourcing the creation of image segmentation algorithms for connectomics. *Frontiers in neuroanatomy*, *9*, p.152591.